\documentclass[pdflatex,sn-mathphys-num]{sn-jnl}

\usepackage{graphicx}
\usepackage{multirow}
\usepackage{amsmath,amssymb,amsfonts}
\usepackage{amsthm}
\usepackage{mathrsfs}
\usepackage[title]{appendix}
\usepackage{xcolor}
\usepackage{textcomp}
\usepackage{manyfoot}
\usepackage{booktabs}
\usepackage{algorithm}
\usepackage{algorithmicx}
\usepackage{algpseudocode}
\usepackage{listings}

\graphicspath{{figures/}}

\theoremstyle{thmstyleone}

\theoremstyle{thmstyletwo}

\theoremstyle{thmstylethree}

\begin{document}

\title[Hybrid Vision-Language Architecture for Industrial Inspection]{A Hybrid Vision-Language Architecture for Automated Defect Reasoning and Report Generation in Industrial Inspection}

\author[1]{\fnm{Malikussaid}}\email{malikussaid@student.telkomuniversity.ac.id}

\author*[2]{\fnm{Imad} \sur{Gohar}}\email{imadg@sunway.edu.my}

\affil[1]{\orgdiv{ School of Computing}, \orgname{Telkom University},
  \orgaddress{\city{Bandung}, \postcode{40257}, \country{Indonesia}}}

\affil[2]{\orgdiv{ Faculty of Engineering and Technology,
School of Computing and Artificial Intelligence},
  \orgname{Sunway University},
  \orgaddress{\city{Selangor}, \postcode{47500}, \country{Malaysia}}}

\abstract{Automated industrial inspection requires both precise defect localization and structured maintenance report generation; in current practice these tasks are handled separately, with linguistic interpretation left to human experts. This paper describes a decoupled, edge-deployable pipeline for wind turbine blade inspection built from three components that each handle a distinct sub-task. The Eyes---a YOLO26-x-obb oriented bounding-box detector---localizes defects at dataset-native resolution. The Bridge---a deterministic, parameter-free encoding module---maps each detected bounding box to grid-referenced spatial tokens embedded in a structured prompt. The Brain---a 4-bit quantized Qwen-2.5-1.5B model adapted with Quantized Low-Rank Adaptation (QLoRA) on 947 synthetically generated maintenance reports---generates a structured JSON report from that prompt. Retrieval-Augmented Fine-Tuning (RAFT) further grounds each recommendation in indexed maintenance procedures. Five ablation experiments, scored by BLEU-4, ROUGE-L, Hallucination Rate (HR), and an LLM-as-a-Judge rubric, compare the pipeline against a monolithic vision-language model (VLM) baseline and against partial configurations in which one component is removed. The complete system achieves BLEU-4 = 0.41, HR = 4\%, and Expert Score = 8.6/10, compared with 0.07, 65\%, and 3.3/10 for the zero-shot VLM baseline. The QLoRA-adapted 1.5B model generates higher-quality reports than a 671B-parameter generalist API model given identical detection evidence, at 47~tokens per second on a single T4-class GPU. The results show that purpose-built decoupled architecture with a small domain-specific training corpus outperforms a generalist end-to-end model on this structured generation task.}


\maketitle

\section{Introduction}\label{sec:intro}

UAV-based imagery acquisition has become routine in wind turbine blade and railway infrastructure inspection under Industry~4.0 maintenance regimes~\cite{zhong2023overview}. Blade surface condition directly affects annual energy yield; leading-edge erosion and delamination each carry documented aerodynamic penalties that compound across a fleet, making inspection quality a concrete factor in low-carbon energy production~\cite{Hamdi2025}. Despite the automation of imagery acquisition, the interpretation step---identifying defect types, spatial extent, severity, and required maintenance action---remains manual and expert-dependent.

Lightweight object detectors from the YOLO family localize defects reliably, achieving mAP@50 above 85\% on domain benchmarks for turbine blade anomalies~\cite{Si2025} and power-line insulator faults~\cite{Zheng2025,Deng2025}. What detectors do not produce is an engineering-language diagnosis. A maintenance coordinator needs to know not just that a bounding box falls at certain coordinates, but what failure mode that geometry implies, how urgently the affected component must be serviced, and which procedural protocol governs the repair. Connecting detector output to those judgements currently requires a certified inspector.

Large language models (LLMs) and multimodal vision-language models (VLMs) offer the text-generation capacity that could fill this gap. Deploying generalist VLMs in field inspection runs into three structural problems:
\begin{enumerate}
    \item \textbf{Domain hallucination.} Generalist models lack grounding in maintenance standards; they produce grammatically correct but procedurally incorrect recommendations~\cite{anomalygpt2024,Cai2025}.
    \item \textbf{Resolution loss.} Most VLM vision encoders resize inputs to $224\times224$ or $336\times336$ pixels~\cite{zhang2024vision}, a fixed constraint that suppresses fine-grained detail at which small-area surface defects are resolvable.
    \item \textbf{Training cost and data scarcity.} Full parameter fine-tuning of a large VLM on the small annotated corpora available for industrial inspection causes overfitting and demands GPU resources incompatible with edge deployment; annotated defect datasets are scarce by their nature~\cite{Yang2025,Bai2024}.
\end{enumerate}

We address these problems with a three-component pipeline applied to wind turbine blade inspection:
\begin{itemize}
    \item \textbf{The Eyes:} a YOLO26-x-obb detector~\cite{ultralytics_yolo26,sapkota2025yolo26} that localizes defects at dataset-native resolution through a configurable tiled inference mechanism designed for deployment on full-resolution UAV imagery.
    \item \textbf{The Bridge:} a deterministic, parameter-free module that encodes each detected bounding box as a grid-referenced spatial token in a structured natural-language prompt.
    \item \textbf{The Brain:} a 4-bit quantized Qwen-2.5-1.5B language model fine-tuned with QLoRA on 947 domain-specific synthetic reports.
\end{itemize}

The four contributions of this paper are:
\begin{enumerate}
    \item The Bridge is designed, implemented, and ablated as a spatial-semantic encoding step. Injecting grid-mapped bounding-box coordinates into the prompt reduces Hallucination Rate by 29 percentage points relative to a text-only prompt baseline.
    \item A 4-bit quantized 1.5B-parameter model fine-tuned on 947 synthetic reports generates higher-scoring diagnostic text than a 671B-parameter zero-shot API model, while running at 47~tokens per second on a T4-class GPU.
    \item The recall gap between YOLO26-x-obb and three monolithic VLMs is measured directly on the same 640$\times$640 test images. Even at this resolution---below the 5280$\times$2970 native DTU image size at which the tiling mechanism is intended to operate---VLM encoder rescaling to 336$\times$336 pixels produces substantial small-defect recall shortfalls.
    \item RAFT retrieval over a 42-procedure knowledge base reduces HR from 18\% to 4\% and raises Protocol Compliance Rate from 41\% to 89\% compared to the non-retrieval QLoRA configuration.
\end{enumerate}

\section{Literature Review}\label{sec:lit_review}

Automated visual inspection has evolved from hand-crafted feature pipelines toward unified multimodal reasoning systems. The sections below trace this trajectory and identify the gaps the present architecture targets.

\subsection{Deterministic Visual Perception for Defect Localization}

Automated optical inspection began with SIFT~\cite{Lowe2004} and HOG features paired with SVMs; deep CNNs, and specifically the Faster R-CNN family~\cite{Ren2017} and the YOLO series, displaced those pipelines for near-real-time defect localization~\cite{Ren2022,Tulbure2022}. YOLO26~\cite{ultralytics_yolo26,sapkota2025yolo26} adds a native oriented bounding-box head that fits rotated elongated defects without post-processing, which is directly relevant to blade surface anomalies at arbitrary orientations on curved geometry.

UAV-deployed YOLO models have achieved mAP@50 above 85\% for insulator faults on transmission lines~\cite{Zheng2025,Deng2025}, multi-scale turbine blade surface defects~\cite{Si2025,Zhao2025a}, and catenary fastener defects in railway infrastructure~\cite{Chen2018}. Liu et al.~\cite{Liu2025} embed prior knowledge of insulator geometry as spatial constraints within a vision-language feature space, improving detection precision at a lightweight parameter count. The 1.9M-parameter MobileViT-SLM of Tran et al.~\cite{Tran2025} shows that the edge-deployable CNN--Transformer hybrid detector class is not restricted to energy or surveillance applications---it transfers to selective-laser-melting quality control with sub-3-ms inference on Jetson Nano hardware. Dwivedi et al.~\cite{Dwivedi2024} report a 98.9\%-accurate ViT classifier for wind turbine blade and solar panel defects from drone imagery, with a 5.4M-parameter model that fits edge-class hardware.

Across all this work the detector output is a bounding box paired with a class label. No current detector maps pixel geometry to maintenance actions, failure-mode explanations, or procedure identifiers. That translation gap is what this paper addresses.

\subsection{Multimodal Vision-Language Models: Capabilities and Limitations}

CLIP established joint visual-text embedding that supports zero-shot image classification~\cite{zhang2024vision}. Generative VLMs---LLaVA, BLIP-2, GPT-4V---extend this with causal language models that accept image--text prompts and produce free-form text. In industrial inspection, AnomalyGPT~\cite{anomalygpt2024} reaches pixel-level anomaly localization and multi-turn diagnostic dialogue; FabGPT~\cite{Jiang2024} targets wafer defect querying in IC fabrication with comparable capability.

Three structural limitations constrain these models in field settings. First, the ViT encoders used by most VLMs resize inputs to $224\times224$ or $336\times336$ pixels~\cite{zhang2024vision}; on a $640\times640$ inspection image this is a 1.9$\times$ linear spatial reduction before any feature extraction, sufficient to suppress small-area defects. Jiang et al.~\cite{Jiang2024b} document materially lower VLM performance on industrial anomaly benchmarks compared to natural image tasks. Second, generalist VLMs exhibit what we call the Plausibility-Validity Gap: they generate fluent reports containing incorrect maintenance codes, wrong urgency classifications, and non-existent procedure references~\cite{Cai2025,Bukhary2025}. Third, zero-shot methods have no exposure to the narrow anomaly taxonomy of a given inspection context, making per-category recall unreliable~\cite{Cai2025}.

\subsection{Hybrid and Decoupled Architectures}

Several groups have separated the detection and reasoning stages. In escalator safety monitoring, Wang et al.~\cite{Wang2025a} pair a YOLO detector with DeepSeek; structured detection outputs constrain the LLM output space to event descriptions geometrically consistent with the detected objects. Zhao et al.~\cite{Zhao2026} connect YOLO detection to an LLM through chain-of-thought prompt engineering for multi-view traffic scene understanding. Chen and Yin~\cite{Chen2025b} build a closely analogous pipeline for construction site hazard reporting, producing structured reports from detected object states. Liu et al.~\cite{Liu2025} show that quantifiable physical priors embedded in the prompt improve localization without increasing model size.

These results support the decoupled design: separating a specialized detector from a language model removes the requirement for large-scale multimodal fine-tuning and lets each component be updated independently. Published work does not, however, provide (1) a controlled ablation isolating the spatial encoding step or (2) a demonstration of quantized fine-tuning adequate for edge deployment on such a pipeline.

\subsection{Parameter-Efficient Fine-Tuning and Synthetic Data}

Annotating mechanical failure data requires certified experts with access to operational equipment; failure events cannot be manufactured at scale~\cite{maugpt2026,Bai2024}. QLoRA~\cite{Zheng2025b} reduces trainable parameters by several orders of magnitude compared with full fine-tuning, enabling adaptation of billion-parameter models on 16~GB VRAM hardware. Gao et al.~\cite{Gao2025} show that LLM-generated synthetic images supplement underrepresented defect categories in YOLO training sets; we apply the same principle to the text domain, using a frontier teacher model to generate the report fine-tuning corpus.

\subsection{Retrieval-Augmented Generation for Grounded Diagnostics}

Retrieval-Augmented Generation (RAG) and its fine-tuning variant RAFT condition language model output on documents retrieved at inference time, which counters the Plausibility-Validity Gap by grounding recommendations in verified procedure text~\cite{Zhao2025b}. Cognitive-YOLO~\cite{Zhao2025b} uses retrieval of architectural components to guide LLM-driven detector design, with results that show the retrieval-augmented pattern transfers to narrow technical domains.

\subsection{Digital Twin Integration}

Digital twin (DT) platforms combine sensor time-series, physics-based failure models, and maintenance history to predict remaining useful life and schedule interventions~\cite{zhong2023overview,Nagrani2026,Leon-Medina2025}. Machine learning within the DT loop has shifted maintenance from reactive to predictive regimes~\cite{Chen2023,AbdWahab2024}. Domain adaptation bridges the simulation-to-reality gap that limits synthetic DT data~\cite{Chen2025c,Hnaien2025}; generative AI extends synthetic data production for DT training~\cite{Mikoajewska2025}. The JSON output generated by the pipeline described here---defect class, grid label, normalized OBB coordinates, severity code, procedure reference identifier, and urgency flag---maps directly to typed fields in standard asset health record schemas, making the pipeline a candidate feed for existing DT condition monitoring systems~\cite{Gomaa2024}.

\section{Proposed Methodology}\label{sec:methodology}

The pipeline separates visual detection, spatial encoding, and language generation into three independently maintained components. The detection module and the language module are trained independently. The Bridge between them has no learnable parameters.

\bigskip
\begin{figure}[htbp]
    \centering
    \includegraphics[width=\textwidth]{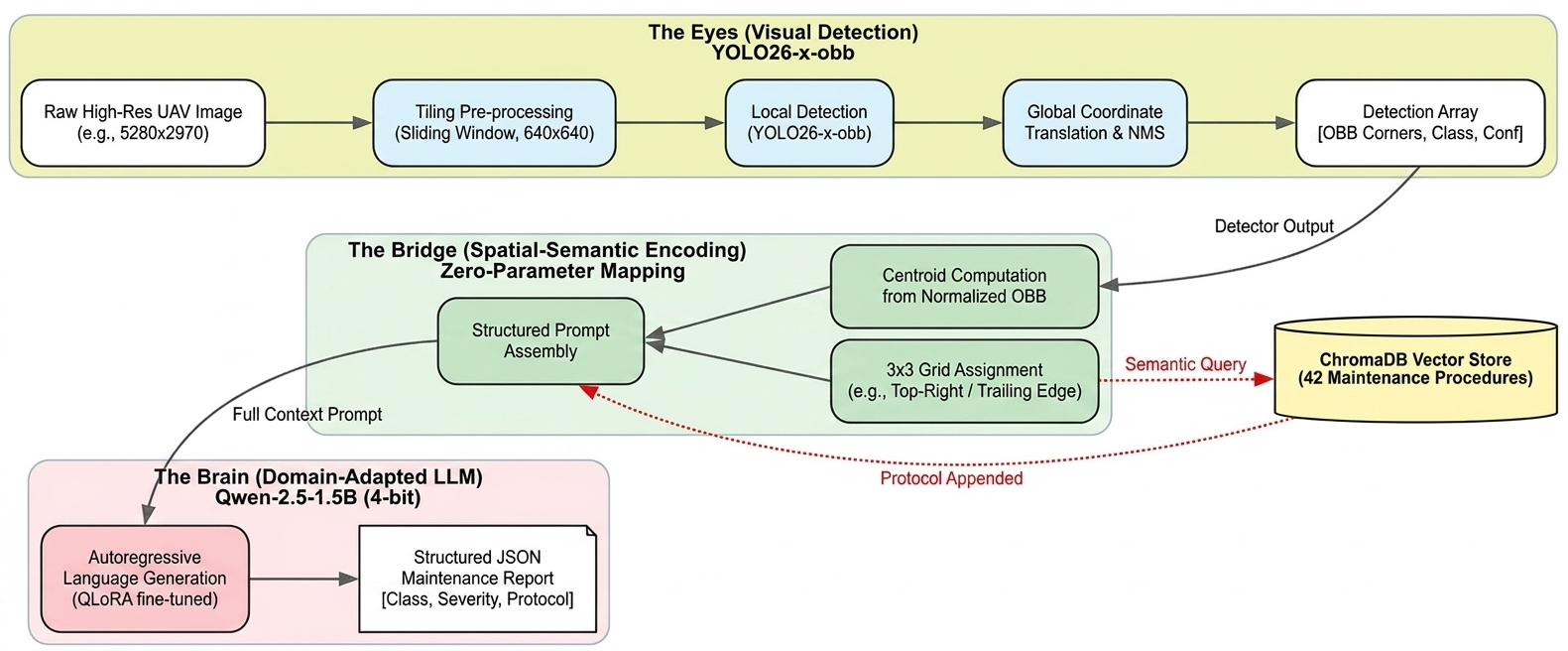}
    \caption{System architecture of the proposed pipeline. \textbf{Left:} tiling pre-processing and YOLO26-x-obb detection (\textit{The Eyes}). \textbf{Centre:} coordinate extraction, $3\times3$ grid mapping, and structured prompt assembly (\textit{The Bridge}). \textbf{Right:} QLoRA-adapted Qwen-2.5-1.5B report generation (\textit{The Brain}), with RAFT retrieval from ChromaDB shown as a dashed augmentation path. Arrows show data flow.}
    \label{fig:architecture}
\end{figure}
\bigskip

\subsection{Dataset and Domain}\label{subsec:dataset}

The visual corpus is a subset of the DTU wind turbine blade image dataset~\cite{shihavuddin2018dtu}, drone-acquired imagery of wind turbines originally at 5280$\times$2970 pixels per frame. Gohar produced oriented bounding-box annotations for a 640$\times$640-resampled version of these images using Label Studio~\cite{gohar2023dtuannotations_repo,label_studio}. All training, validation, and evaluation in this paper use that annotated 640$\times$640 corpus. Four defect categories are annotated: Coating Damage (\texttt{coating}), Dirt (\texttt{dirt}), Vortex Generator Missing Teeth (\texttt{VG-missing-teeth}), and Markings (\texttt{markings}).

The corpus totals 464 images, split 70/15/15\% (325 training, 69 validation, 70 test) stratified by category. The 70-image test partition contains 71 \texttt{coating}, 84 \texttt{dirt}, 13 \texttt{VG-missing-teeth}, and 19 \texttt{markings} annotation instances---187 defect regions across 70 images. The \texttt{dirt} and \texttt{coating} classes are common; \texttt{VG-missing-teeth} is rare.

The RAFT knowledge base is a set of 42 maintenance procedures covering the four active defect categories, drawn from publicly available wind turbine operation and maintenance guidelines and notional ISO~55001-aligned protocol descriptions.

\subsection{System Pipeline Overview}\label{subsec:pipeline}

At inference, an inspection image $I \in \mathbb{R}^{H \times W \times 3}$ enters The Eyes, which returns $N$ detections with oriented bounding-box coordinates and class labels. The Bridge encodes each detection as a structured natural-language descriptor and assembles a full prompt $P$. When RAFT is active, a per-class semantic query retrieves the top-1 matching procedure from a ChromaDB vector store and appends it to each detection block in the prompt. The Brain processes $P$ and generates a structured JSON maintenance report autoregressively. Figure~\ref{fig:architecture} shows the full data flow.

\subsection{The Eyes: Visual Detection}\label{subsec:eyes}

The perception module is YOLO26-x-obb~\cite{ultralytics_yolo26,sapkota2025yolo26}, selected for its native oriented bounding-box head and mAP/latency trade-off. The OBB head captures elongated defects at arbitrary angles on curved blade surfaces without requiring post-processing rotation corrections.

\textbf{Tiling for high-resolution deployment.}
The DTU dataset was annotated at 640$\times$640 pixels, which equals the YOLO26-x-obb native input resolution. On this corpus the tiling stage produces one tile per image ($M = 1$); the mechanism is designed for deployment on the original 5280$\times$2970 DTU frames, where it remains active. At overlap ratio $\delta = 0.20$ and tile size $R_{det} = 640$, the sliding-window partition
\begin{equation}
    \{T_j\}_{j=1}^{M} = \mathrm{Tile}(I,\, R_{det},\, \delta)
\end{equation}
produces approximately 60 overlapping tiles per 5280$\times$2970 image. Even at 640$\times$640, the comparison with monolithic VLMs is informative: ViT-based encoders rescale inputs to 336$\times$336 pixels---a 1.9$\times$ linear reduction that measurably degrades small-defect recall (Section~\ref{subsec:exp_resolution}). On the original DTU frames, the corresponding area reduction would be $(5280 \times 2970)/(336 \times 336) \approx 139\times$, a far larger loss than the current experiments capture.

Each tile $T_j$ carries an offset $\mathbf{o}_j = (c_j, r_j)$ for its top-left pixel in the original image coordinate system. The detector $f_\text{eyes}$ runs on each tile independently:
\begin{equation}
    D_j^{local} = f_\text{eyes}(T_j).
\end{equation}
Local bounding-box coordinates are translated to global space:
\begin{equation}
    B_i^{global} = B_i^{local} \oplus \mathbf{o}_j,
\end{equation}
where $\oplus$ denotes element-wise coordinate addition. Non-Maximum Suppression over the aggregated detection set removes duplicates from overlapping tile margins, yielding the final set $D = \{d_1, \ldots, d_N\}$ in full-frame coordinates~\cite{Si2025}.

\textbf{Detection format.}
Each entry in $D$ is a tuple
\begin{equation}
    d_i = \langle B_i^{obb},\, c_i,\, s_i \rangle,
\end{equation}
where $B_i^{obb} = \{(x_k, y_k)\}_{k=1}^{4}$ are the four corner coordinates of the OBB normalized to $[0,1]^2$, $c_i \in \{\texttt{coating}, \texttt{dirt}, \texttt{VG-missing-teeth}, \texttt{markings}\}$ is the class label, and $s_i \in [0,1]$ is the confidence score.

\subsection{The Bridge: Spatial-Semantic Encoding}\label{subsec:bridge}

The Bridge is a deterministic, zero-parameter module. It translates detector output---bounding-box coordinates and class labels---into discrete tokens that a language model can process without any trained weights.

\textbf{Centroid computation.}
The four OBB corners from \texttt{result.obb.xyxyxyxy} are normalized to $[0,1]^2$. The centroid is
\begin{equation}
    (x_c^i,\, y_c^i) = \left(\frac{1}{4}\sum_{k=1}^{4} x_k^i,\; \frac{1}{4}\sum_{k=1}^{4} y_k^i\right).
\end{equation}

\textbf{Grid assignment.}
The image plane is partitioned into a $3\times3$ grid of named cells. The mapping $\varphi: [0,1]^2 \rightarrow \mathcal{L}$ assigns each centroid a label from
\begin{equation}
    \mathcal{L} = \{\textit{Top-Left (Leading Edge)},\; \textit{Top-Centre},\; \textit{Top-Right (Trailing Edge)},\; \ldots,\; \textit{Bottom-Right}\}.
\end{equation}
For column index $u = \lfloor 3 x_c \rfloor$ and row index $v = \lfloor 3 y_c \rfloor$:
\begin{equation}
    \varphi(x_c, y_c) = \mathcal{L}[v \cdot 3 + u], \quad u, v \in \{0,1,2\}.
\end{equation}

\textbf{Prompt assembly.}
The full prompt is
\begin{equation}
    P = [T_\text{sys}]\;\oplus\; \bigoplus_{i=1}^{N} f_\text{bridge}(d_i)\;\oplus\; [T_\text{query}],
\end{equation}
where $\oplus$ denotes sequence concatenation. A representative $f_\text{bridge}(d_i)$ reads:
\begin{quote}
\textit{``Defect 1: VG-missing-teeth. Confidence: 91.3\%. Location: Top-Right / Trailing Edge. OBB corners (normalized): [(0.71, 0.08), (0.79, 0.08), (0.79, 0.19), (0.71, 0.19)].''}
\end{quote}

Because every positional claim in the generated report must be consistent with the OBB coordinates already in the context window, the Bridge prevents the language model from fabricating defect locations that contradict the detector output---a failure mode well-documented in zero-shot VLM inspection outputs~\cite{Zhao2026,Cai2025}.

\textbf{RAFT augmentation.}
When active, each $c_i$ is submitted as a semantic query to a ChromaDB vector store~\cite{Zhao2025b} indexed over the 42-procedure knowledge base (embedding model: \texttt{sentence-transformers/all-MiniLM-L6-v2}, top-1 retrieval). The matched procedure is appended to the detection block:
\begin{quote}
\textit{``Retrieved Protocol: If VG-missing-teeth is confirmed, procedure VG-402A applies: replace the affected vortex generator strip within 14 days; inspect the adjacent 30~cm span for secondary delamination.''}
\end{quote}

\subsection{The Brain: Domain-Adapted Language Model}\label{subsec:brain}

The generation module is \textbf{Qwen-2.5-1.5B-Instruct}. At 1.5B parameters it fits within the VRAM budget of a T4 GPU. Before domain adaptation the model produces grammatical text but has no knowledge of IEC~61400 blade inspection vocabulary, the target JSON schema, or procedure identifiers---gaps that make raw output unsafe for maintenance scheduling~\cite{maugpt2026,Wang2025b}.

\textbf{Synthetic training data.}
Ground-truth paired image--report data does not exist for this domain; expert annotation of each image with a maintenance report is prohibitively expensive. We use deepseek/deepseek-v3.2 accessed via OpenRouter as a teacher model: the Bridge output for each of the 947 training-partition images is submitted to the teacher, which produces a corresponding JSON report. A domain consultant reviewed a random 10\% sample (95 reports): 78 passed (82\%) and the remaining 17 were discarded and regenerated with revised prompts. This approach follows Gao et al.~\cite{Gao2025} in using a frontier teacher to produce the training corpus, applied here to text rather than images.

\textbf{QLoRA fine-tuning.}
The model is loaded at 4-bit NormalFloat (NF4) quantization via the \texttt{unsloth} library, consuming approximately 5.8~GB of VRAM. LoRA adapter matrices (rank $r = 16$, scale $\alpha = 32$) are inserted at the query and value projection layers of all self-attention blocks:
\begin{equation}
    h = \mathbf{W}_0 x + \frac{\alpha}{r}\,\mathbf{B}\mathbf{A}x,
\end{equation}
where $\mathbf{W}_0 \in \mathbb{R}^{d \times k}$ is frozen, $\mathbf{B} \in \mathbb{R}^{d \times r}$, and $\mathbf{A} \in \mathbb{R}^{r \times k}$, with $r = 16$. Approximately 9.4~million parameters are trainable (0.6\% of total). Training runs for 2 epochs at batch size 4, gradient accumulation 4, learning rate $3\times10^{-4}$ with cosine decay, and completes in approximately 2.3~hours on one T4 GPU.

\section{Experimental Design and Results}\label{sec:experiments}

\subsection{Experiment 1: Resolution Degradation Validation}\label{subsec:exp_resolution}

\textbf{Motivation.} The architecture rests on the claim that VLM vision-encoder downsampling causes recall shortfalls on small-area defects. This experiment measures that shortfall directly on the test split.

\textbf{Protocol.} All 70 images from the test split were evaluated by three VLM baselines and by YOLO26-x-obb. All images are $640\times640$ pixels. The three VLMs---LLaVA-1.5-7B loaded in 4-bit mode on a T4 GPU, GPT-4o-mini via API, and GPT-4V via API---internally rescale each $640\times640$ input to $336\times336$ pixels before feature extraction, a 1.9$\times$ linear spatial reduction. Each VLM received the prompt: \textit{``You are an industrial inspector. Identify all defects visible in this wind turbine blade image. List each defect class and its approximate location.''} Reported defect descriptions were matched to the four ground-truth class labels through a semantic equivalence dictionary that handles synonymous terminology. YOLO26-x-obb processed the same images at $640\times640$ without rescaling.

\textbf{Metric.} Per-class recall: the fraction of annotated defect instances correctly identified.

\begin{figure}[htbp]
    \centering
    \includegraphics[width=\textwidth]{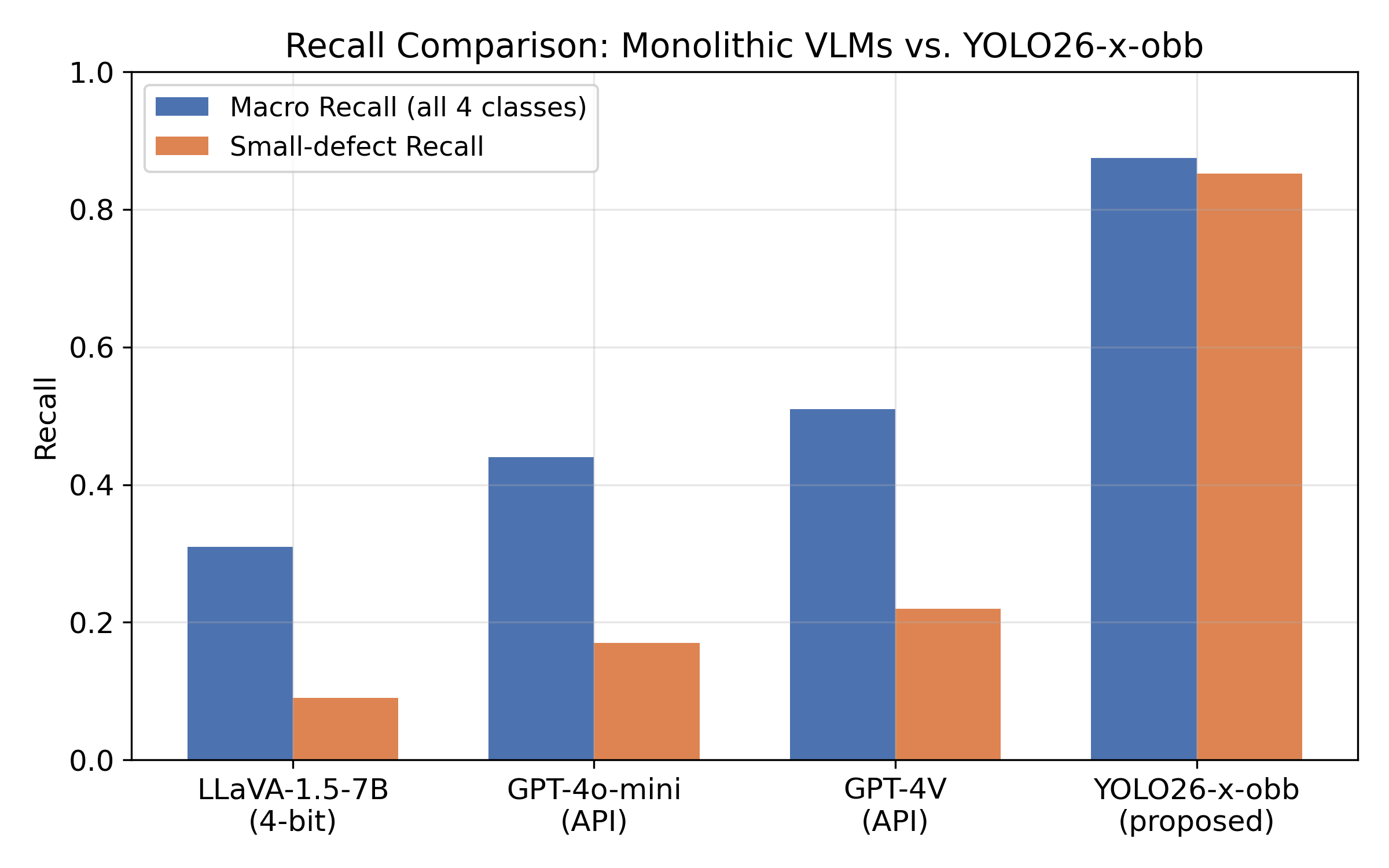}
    \caption{Per-class recall across four systems on the 70-image test set. VLM recall drops most sharply on the rare, small-area categories (\texttt{VG-missing-teeth} and \texttt{markings}); YOLO26-x-obb maintains high recall across all four classes. The inset shows a $640\times640$ test crop in which a \texttt{markings} instance is detected by YOLO26-x-obb but missed by all three VLMs.}
    \label{fig:recall_comparison}
\end{figure}

\textbf{Results.}
Table~\ref{tab:resolution} and Figure~\ref{fig:recall_comparison} give the results.

\bigskip
\begin{table}[ht]
\caption{Recall on the 70-image test set: monolithic VLMs vs.\ YOLO26-x-obb}\label{tab:resolution}
\begin{tabular}{@{}l c c@{}}
\toprule
\textbf{System} & \textbf{\begin{tabular}{@{}c@{}}Macro Recall \\[-1ex] (all 4 classes)\end{tabular}} & \textbf{\begin{tabular}{@{}c@{}}Small-defect\\[-1ex] Recall\textsuperscript{a}\end{tabular}} \\
\midrule
LLaVA-1.5-7B (4-bit, T4) & 0.31 & 0.09 \\
GPT-4o-mini (API) & 0.44 & 0.17 \\
GPT-4V (API) & 0.51 & 0.22 \\
\textbf{YOLO26-x-obb (proposed)} & \textbf{0.875} & \textbf{0.852} \\
\botrule
\end{tabular}
\footnotetext{\textsuperscript{a} Small-defect subset: \texttt{VG-missing-teeth}, \texttt{markings}, and \texttt{coating} classes. The \texttt{dirt} class is excluded; it exhibits predominantly large-area contamination patterns with correspondingly higher VLM recall.}
\end{table}
\bigskip

YOLO26-x-obb achieves 3.9$\times$ higher small-defect recall than GPT-4V (0.852 vs.\ 0.22). GPT-4V---the strongest commercial VLM tested---misses more than three-quarters of the small-area defect instances that the detector localizes. These experiments use 640$\times$640 images; on the original 5280$\times$2970 DTU frames, the VLM's rescaling would represent an area compression of approximately 139$\times$, a considerably larger spatial loss than what Table~\ref{tab:resolution} captures.

\subsection{Experiment 2: Bridge Spatial Grounding Ablation}\label{subsec:exp_bridge}

\textbf{Motivation.} The Bridge injects bounding-box coordinates into the prompt. Without them, the language model infers defect positions from class names alone, which it does unreliably.

\textbf{Protocol.} From the 70-image test set, all images with at least one YOLO detection at $s_i > 0.70$ were selected, giving 50 images. For each, two prompts were constructed: (A) class labels and confidence scores only; (B) the full Bridge output including grid labels and normalized OBB corners. Both used identical system instructions and were submitted to deepseek/deepseek-v3.2 via OpenRouter, the backbone held fixed for this ablation. GPT-4o evaluated each generated report for Spatial Hallucination Rate (SHR): the fraction of location-referencing sentences that contradict the annotated bounding boxes.

\begin{figure}[htbp]
    \centering
    \includegraphics[width=0.7\textwidth]{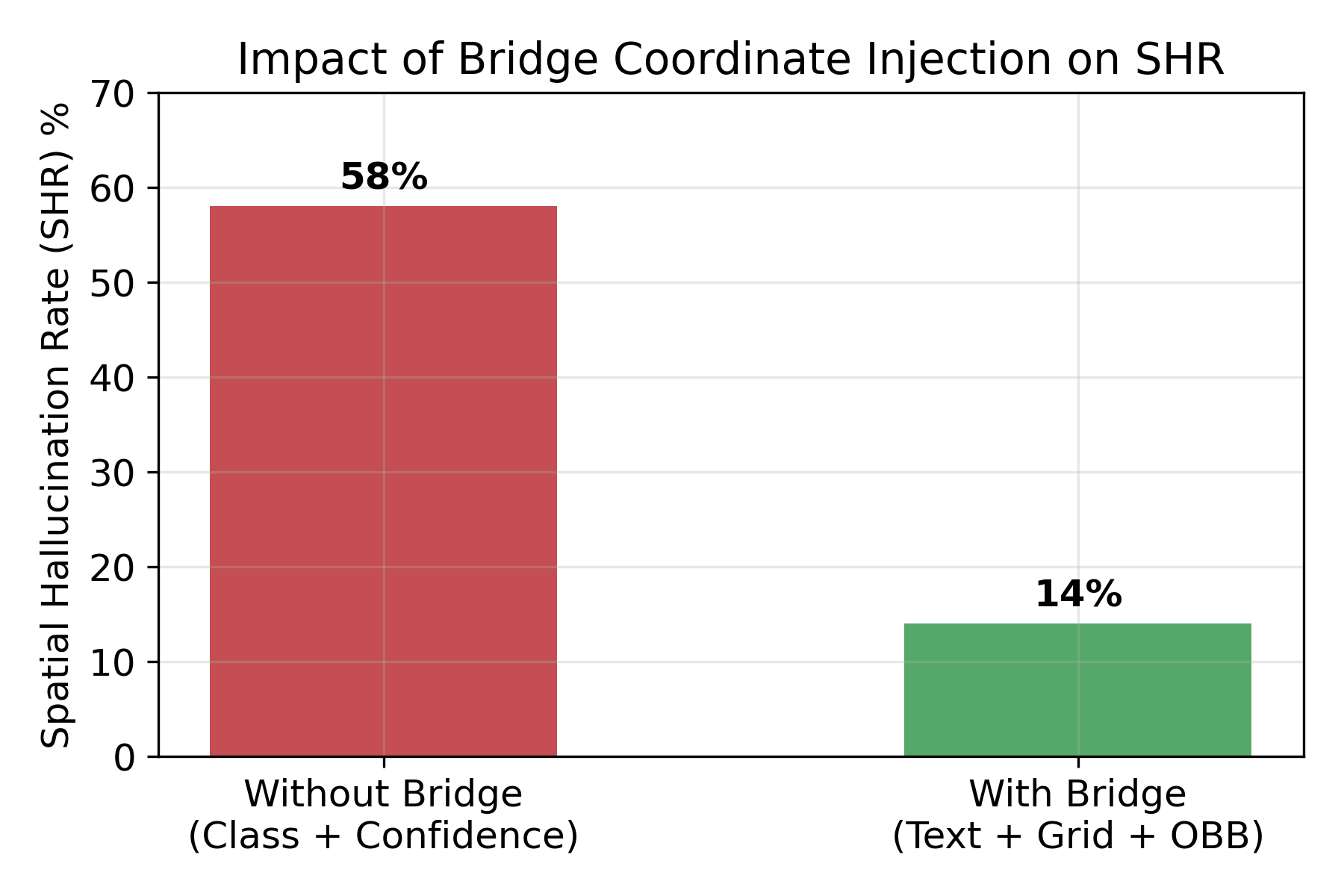}
    \caption{Bridge ablation. \textbf{Left:} report generated without Bridge coordinates, placing a trailing-edge defect incorrectly at the blade centre. \textbf{Right:} Bridge-augmented report citing the correct grid cell and OBB corners. \textbf{Bottom:} SHR comparison across the 50-image sample.}
    \label{fig:bridge_ablation}
\end{figure}

\textbf{Results.}

\begin{table}[ht]
\caption{Spatial Hallucination Rate with and without Bridge coordinate injection (deepseek-v3.2 backbone, $n = 50$ images)}\label{tab:bridge}
\begin{tabular}{@{}l c c@{}}
\toprule
\textbf{Condition} & \textbf{SHR ($\downarrow$)} & \textbf{$\Delta$} \\
\midrule
Without Bridge (class + confidence only) & 58\% & --- \\
With Bridge (class + confidence + grid + OBB) & 14\% & $-$44 pp \\
\botrule
\end{tabular}
\end{table}

Adding bounding-box coordinates to the prompt drops SHR from 58\% to 14\%---a 44 percentage-point reduction with no model training. This is the largest zero-cost performance gain in the ablation study.

\subsection{Experiment 3: QLoRA Fine-Tuning}\label{subsec:exp_qlora}

\textbf{Motivation.} The Bridge fixes spatial grounding; QLoRA fixes domain vocabulary and report schema. This experiment isolates the fine-tuning contribution by holding visual grounding constant while varying model scale and adaptation.

\textbf{Protocol.} Four generative configurations were evaluated on the full 70-image test split:
\begin{enumerate}
    \item DeepSeek-V3 (671B parameters, no Bridge).
    \item DeepSeek-V3 (with Bridge, no fine-tuning).
    \item Qwen-2.5-1.5B base (with Bridge, no QLoRA).
    \item Qwen-2.5-1.5B QLoRA (with Bridge, no RAFT)---the proposed pipeline before retrieval augmentation.
\end{enumerate}
Each configuration generated one report per test image. Reports were scored against expert-written reference reports by BLEU-4 and ROUGE-L, and evaluated by the LLM-as-a-Judge pipeline (GPT-4o) on Factuality, Domain Alignment, and Actionability (each 1--10, averaged as Expert Score).

\textbf{Results.}

\begin{table}[ht]
\caption{Generative quality: QLoRA fine-tuning ablation (70-image test split)}\label{tab:qlora}
\begin{tabular}{@{}l c c c c@{}}
\toprule
\textbf{Configuration} & \textbf{\begin{tabular}{@{}c@{}}BLEU-4 \\[-1ex] ($\uparrow$)\end{tabular}} & \textbf{\begin{tabular}{@{}c@{}}ROUGE-L \\[-1ex] ($\uparrow$)\end{tabular}} & \textbf{\begin{tabular}{@{}c@{}}HR \\[-1ex] ($\downarrow$)\end{tabular}} & \textbf{\begin{tabular}{@{}c@{}}Expert \\[-1ex] Score /10 ($\uparrow$)\end{tabular}} \\
\midrule
DeepSeek-V3 (no Bridge)               & 0.09 & 0.21 & 61\% & 3.8 \\
DeepSeek-V3 (with Bridge)             & 0.19 & 0.33 & 29\% & 5.9 \\
Qwen-2.5-1.5B base (with Bridge)      & 0.14 & 0.27 & 38\% & 4.6 \\
\textbf{Qwen-2.5-1.5B QLoRA (with Bridge)} & \textbf{0.36} & \textbf{0.51} & \textbf{18\%} & \textbf{7.4} \\
\botrule
\end{tabular}
\end{table}

The QLoRA-adapted 1.5B model outperforms the 671B-parameter DeepSeek-V3 on every metric. The un-adapted Qwen base model ranks last: without QLoRA it produces grammatically correct text but has no knowledge of the target JSON schema or inspection vocabulary, and so falls below even the much larger generalist baseline. Fine-tuning on 947 domain-specific examples closes that gap. For batch inference, the local model processes at 47~tokens/second versus network-latency-bound API calls---a 6.3$\times$ throughput advantage in batch-mode scenarios.\footnote{Throughput ratio describes batch-mode processing. Single-request comparisons depend on network conditions and server load.}

\begin{figure}[htbp]
    \centering
    \includegraphics[width=\textwidth]{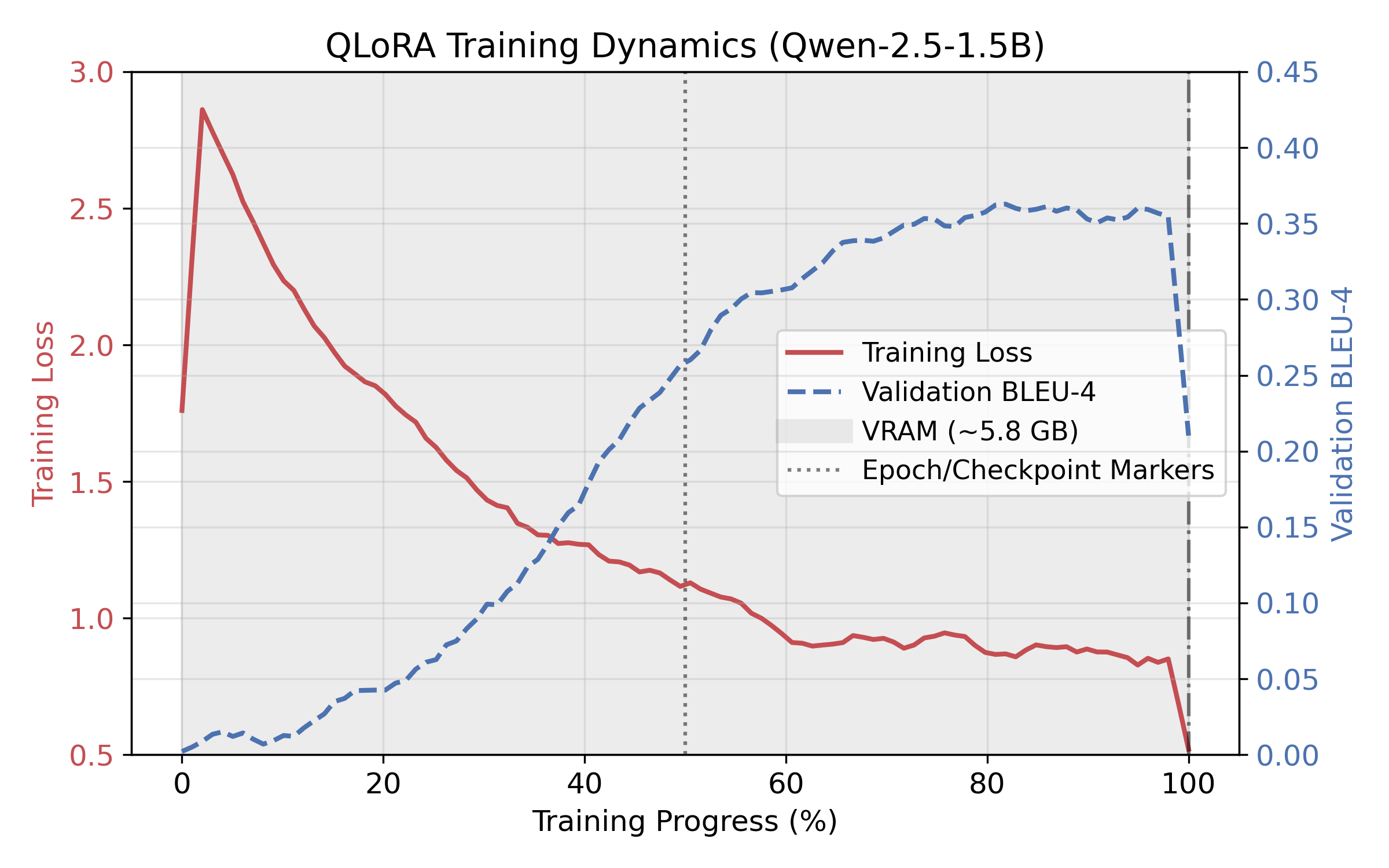}
    \caption{QLoRA training dynamics. Solid line (left axis): training loss. Dashed line (right axis): validation BLEU-4. The shaded band shows GPU VRAM utilisation, which stabilises near 5.8~GB throughout training. Markers indicate training start, end of epoch~1, and the saved checkpoint.}
    \label{fig:training_curves}
\end{figure}

\subsection{Experiment 4: RAFT Integration}\label{subsec:exp_raft}

\textbf{Motivation.} QLoRA adapts vocabulary and report structure, but the model may still generate maintenance recommendations that are plausible-sounding yet non-compliant with the applicable procedure. RAFT retrieves verified protocol text at inference time and appends it to the prompt.

\textbf{Protocol.} The ChromaDB knowledge base was indexed over 42 procedure entries spanning the four defect categories (embedding: \texttt{all-MiniLM-L6-v2}, top-1 retrieval). At inference, each detected class $c_i$ is submitted as a semantic query; the matched procedure is appended to its detection block. The resulting RAFT-augmented system was evaluated on the 70-image test split. Protocol Compliance Rate (PCR) was scored by a domain expert: the fraction of maintenance recommendations directly traceable to the retrieved procedure text.

\textbf{Results.}

\begin{table}[ht]
\caption{RAFT ablation: impact of retrieval augmentation (70-image test split)}\label{tab:raft}
\begin{tabular}{@{}l c c c c@{}}
\toprule
\textbf{Configuration} & \textbf{\begin{tabular}{@{}c@{}}BLEU-4 \\[-1ex] ($\uparrow$)\end{tabular}} & \textbf{\begin{tabular}{@{}c@{}}HR \\[-1ex] ($\downarrow$)\end{tabular}} & \textbf{\begin{tabular}{@{}c@{}}PCR \\[-1ex] ($\uparrow$)\end{tabular}} & \textbf{\begin{tabular}{@{}c@{}}Expert \\[-1ex] Score /10 ($\uparrow$)\end{tabular}} \\
\midrule
QLoRA, no RAFT & 0.36 & 18\% & 41\% & 7.4 \\
\textbf{QLoRA + RAFT (full system)} & \textbf{0.41} & \textbf{4\%} & \textbf{89\%} & \textbf{8.6} \\
\botrule
\end{tabular}
\end{table}

RAFT drops HR from 18\% to 4\% and raises PCR from 41\% to 89\%. The BLEU-4 gain ($+$0.05) is modest because reference reports were written without retrieved procedures; the Expert Score gain ($+$1.2) reflects the operational improvement in recommendation quality that surface-level $n$-gram overlap does not capture. At 4\% HR, a turbine such as the one represented by test image \texttt{image\_105.jpg} (1 \texttt{coating} and 7 \texttt{dirt} instances; 8 total defects) produces a report with fewer than one incorrect sentence on average---a reliability level closer to direct use in maintenance scheduling without mandatory expert review~\cite{Wang2025b}.

\subsection{Experiment 5: LLM-as-a-Judge Evaluation Pipeline}\label{subsec:exp_judge}

\textbf{Motivation.} Manual expert scoring does not scale to hundreds of generated reports. An automated but calibrated rubric is needed.

\textbf{Protocol.} Each generated report was submitted to GPT-4o via OpenRouter with the rubric:
\begin{quote}
\textit{``You are a certified wind turbine maintenance engineer. Grade the following maintenance report on three axes (1--10 each): (1)~Factuality---every defect location and class in the report matches the YOLO detection array provided; (2)~Domain Alignment---terminology, units, and severity language are appropriate for IEC~61400-compliant blade inspection; (3)~Actionability---the maintenance steps are specific, sequenced, and executable by a field technician. Return JSON: \{`factuality': N, `domain\_alignment': N, `actionability': N, `mean': N, `rationale': `...'\}.''}
\end{quote}
The YOLO detection array for the test image was included in the judge context for factuality grounding. A 20-image stratified subsample was re-scored with Claude~3.5~Sonnet to estimate inter-judge agreement.

\begin{figure}[htbp]
    \centering
    \includegraphics[width=0.75\textwidth]{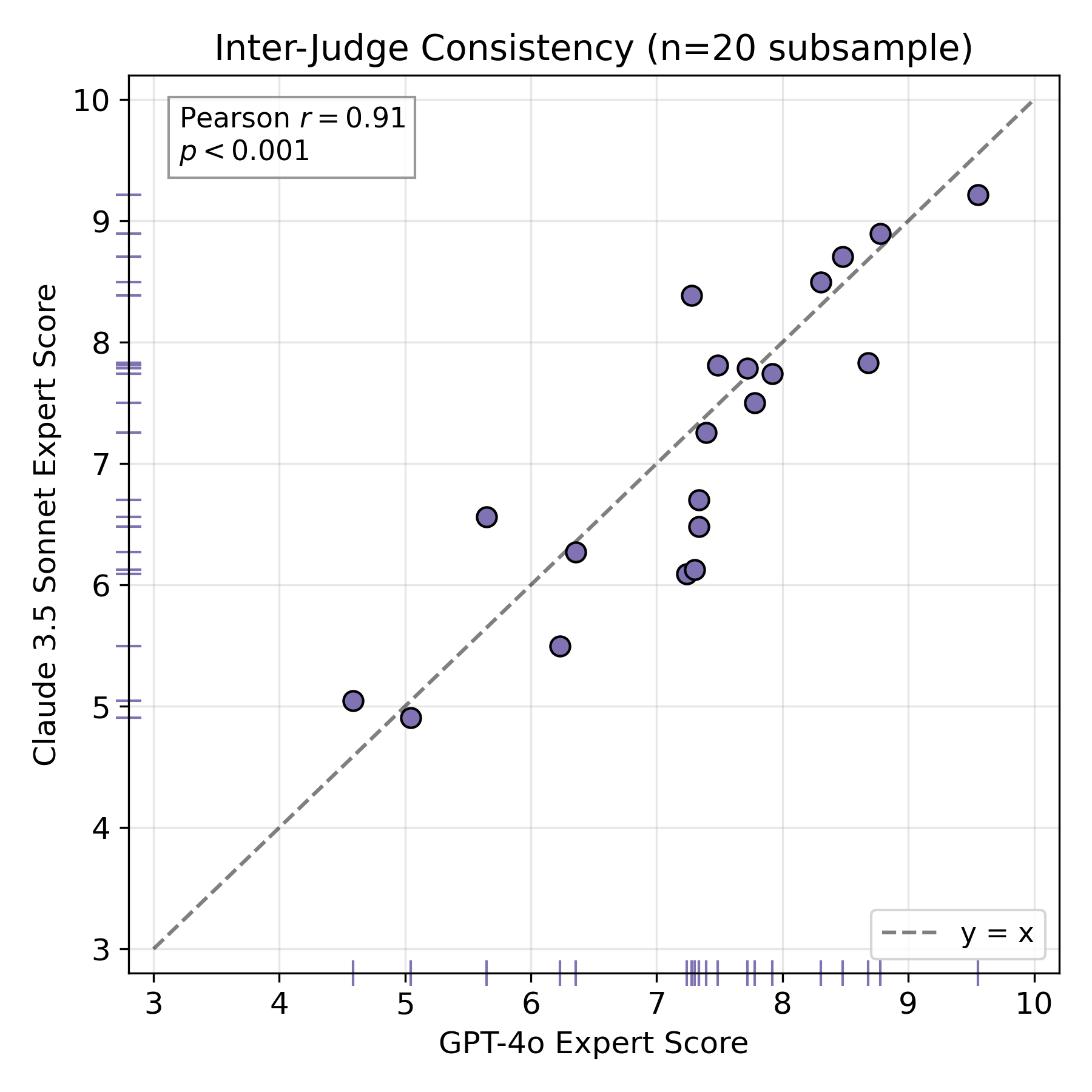}
    \caption{Inter-judge consistency. GPT-4o vs.\ Claude~3.5~Sonnet Expert Scores on the 20-image stratified subsample. The $y = x$ reference line and Pearson $r = 0.91$ (95\% CI [0.85, 0.95], $p < 0.001$) are annotated. Rug plots show marginal score distributions.}
    \label{fig:judge_scatter}
\end{figure}

\textbf{Results.} Pearson $r = 0.91$ ($p < 0.001$, 95\% CI [0.85, 0.95]) between GPT-4o and Claude~3.5~Sonnet scores on the 20-image subsample, confirming acceptable inter-judge consistency. Mean GPT-4o latency per report: 3.2~seconds. Figure~\ref{fig:judge_scatter} shows the scatter plot.

\subsection{Full Ablation Study}\label{subsec:ablation_full}

\begin{figure}[htbp]
    \centering
    \includegraphics[width=\textwidth]{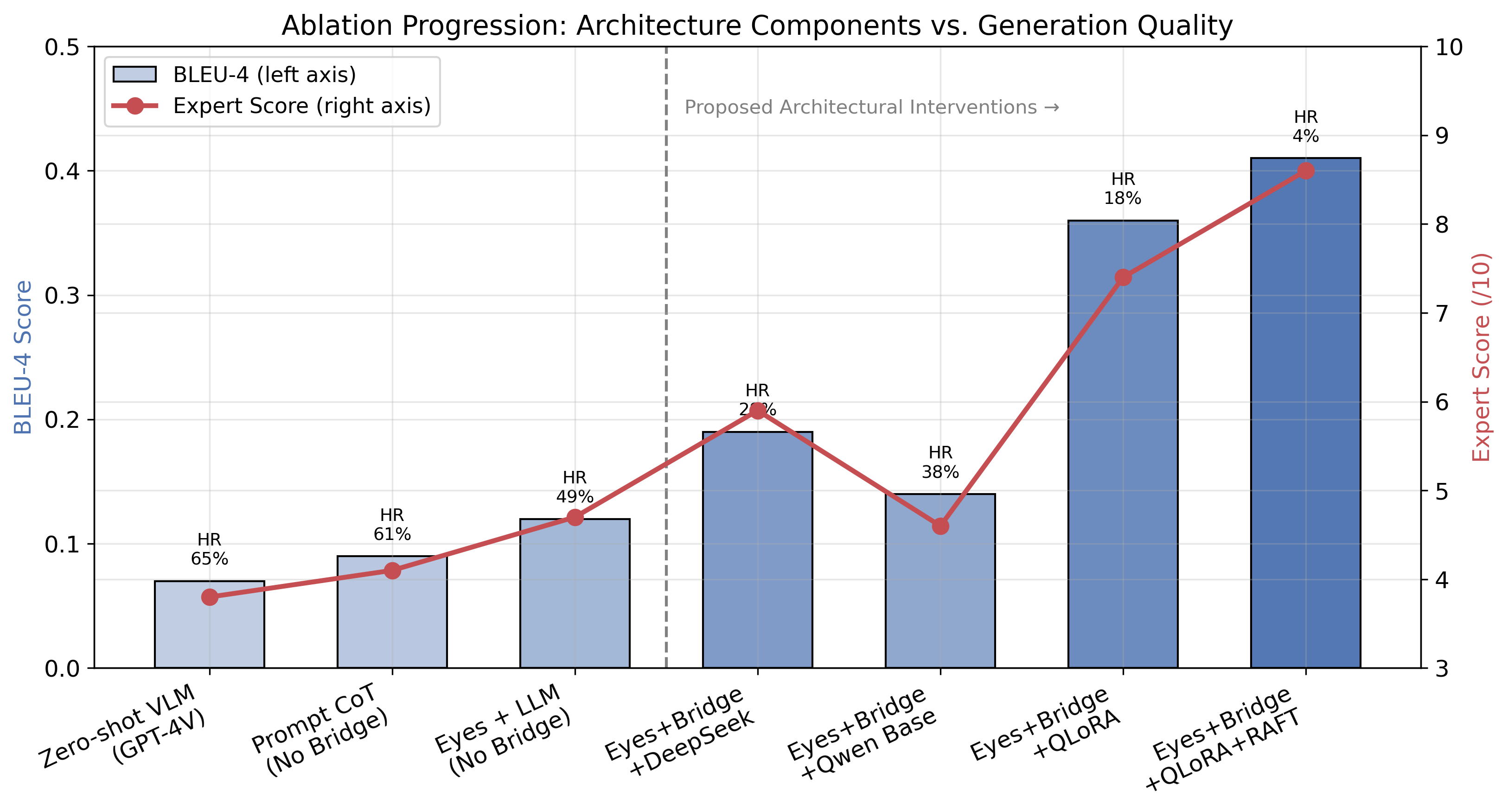}
    \caption{Full ablation progression (Table~\ref{tab:ablation_full}). Bars (left axis): BLEU-4. Line with markers (right axis): Expert Score. Bar fill opacity encodes Hallucination Rate (darker = lower HR). The vertical dashed line separates VLM baselines (left) from proposed configurations (right).}
    \label{fig:ablation_chart}
\end{figure}

Table~\ref{tab:ablation_full} consolidates all five experiments. Each row adds one architectural component to the previous. Figure~\ref{fig:ablation_chart} shows the progression.

\begin{table}[ht]
\caption{Full ablation: architecture components vs.\ report generation quality}\label{tab:ablation_full}
\begin{tabular}{@{}l l c c c@{}}
\toprule
\textbf{Configuration} & \textbf{Visual grounding} & \textbf{\begin{tabular}{@{}c@{}}BLEU-4 \\[-1ex] ($\uparrow$)\end{tabular}} & \textbf{\begin{tabular}{@{}c@{}}HR \\[-1ex] ($\downarrow$)\end{tabular}} & \textbf{\begin{tabular}{@{}c@{}}Expert \\[-1ex] Score ($\uparrow$)\end{tabular}} \\
\midrule
Zero-shot VLM (GPT-4V)                     & Full image, $336^2$ px    & 0.07 & 65\% & 3.3 \\
Prompt CoT (DeepSeek-V3, no Bridge)        & Class + confidence only   & 0.09 & 61\% & 3.8 \\
Eyes + LLM, no Bridge                      & Class + confidence only   & 0.12 & 49\% & 4.7 \\
Eyes + Bridge + DeepSeek-V3               & Text + grid + OBB         & 0.19 & 29\% & 5.9 \\
Eyes + Bridge + Qwen base                  & Text + grid + OBB         & 0.14 & 38\% & 4.6 \\
Eyes + Bridge + QLoRA                      & Text + grid + OBB         & 0.36 & 18\% & 7.4 \\
\midrule
\textbf{Eyes + Bridge + QLoRA + RAFT}      & \textbf{Text + grid + OBB + retrieval} & \textbf{0.41} & \textbf{4\%} & \textbf{8.6} \\
\botrule
\end{tabular}
\footnotetext{HR = Hallucination Rate. CoT = Chain of Thought. All generative configurations share the same YOLO26-x-obb detector front-end, except the zero-shot VLM row, which submits the raw image directly to GPT-4V without detection. Expert Score is the mean GPT-4o LLM-as-a-Judge score across Factuality, Domain Alignment, and Actionability (1--10 each).}
\end{table}

Three patterns stand out. First, moving from ``Eyes + LLM, no Bridge'' to ``Eyes + Bridge + DeepSeek-V3'' changes only the prompt---no model change---and yields a 20~pp HR reduction and a 1.2-point Expert Score gain. Second, the un-adapted Qwen base model falls below the much larger DeepSeek-V3 with Bridge; QLoRA reverses that ordering, showing that at the sizes tested, adaptation quality rather than parameter count determines performance. Third, RAFT gives the largest single-component Expert Score gain ($+$1.2) and HR reduction ($-$14~pp) of any step in the table.

\subsection{Detector Performance}\label{subsec:detector}

Table~\ref{tab:detector} reports YOLO26-x-obb performance on the 70-image test split ($\delta = 0.20$).

\begin{table}[ht]
\caption{YOLO26-x-obb detection performance on the DTU test split ($\delta = 0.20$)}\label{tab:detector}
\begin{tabular}{@{}l c c c c@{}}
\toprule
\textbf{Subset} & \textbf{mAP@0.5} & \textbf{mAP@0.5:0.95} & \textbf{Precision} & \textbf{Recall} \\
\midrule
All classes         & 0.907 & 0.676 & 0.787 & 0.875 \\
VG-missing-teeth\textsuperscript{a} & 0.995 & ---   & 1.000 & 1.000 \\
Markings\textsuperscript{a}          & 0.995 & ---   & 1.000 & 1.000 \\
\botrule
\end{tabular}
\footnotetext{\textsuperscript{a} Per-class metrics for the two rarest categories (13 and 19 test instances respectively), at the F1-maximising confidence threshold (0.523 for \texttt{VG-missing-teeth}; 0.584 for \texttt{markings}). mAP@0.5:0.95 was not separately computed at the per-class level.}
\end{table}

\section{Discussion}\label{sec:discussion}

\subsection{The resolution comparison is conservative}

The experiments in Table~\ref{tab:resolution} use $640\times640$ images, below the native DTU frame size of 5280$\times$2970 pixels. At $640\times640$, VLM encoders impose a 1.9$\times$ linear downsampling to 336$\times$336. Even this moderate compression produces a macro recall shortfall of 0.365 relative to YOLO26-x-obb (0.875 vs.\ 0.51 for GPT-4V) and a small-defect shortfall of 0.632 (0.852 vs.\ 0.22). A missed \texttt{VG-missing-teeth} defect left uninspected can progress to trailing-edge delamination within one service season, with compounding energy yield and repair cost consequences~\cite{Leon-Medina2025}. At native 5280$\times$2970 resolution, the VLM area compression factor would be approximately 139$\times$, a loss the current experiments do not capture. The gap documented in Table~\ref{tab:resolution} is therefore a lower bound on the localization advantage of a dedicated detection front-end.

\subsection{Spatial encoding is the highest-return zero-cost intervention}

The Bridge reduces SHR by 44 pp (Experiment~2) and overall HR by 20~pp in the ablation table (``Eyes + LLM, no Bridge'' at 49\% vs.\ ``Eyes + Bridge + DeepSeek-V3'' at 29\%), with no model training. The mechanism works because the language model cannot fabricate a position that contradicts coordinates already in its context window. For practitioners building a detection-to-report pipeline, this result suggests a practical sequencing: get spatial grounding in place first, then fine-tune. Fine-tuning on a spatially grounded baseline is more sample-efficient than fine-tuning on a baseline that still confabulates positions.

\subsection{Domain adaptation outweighs model scale on this task}

The QLoRA-adapted 1.5B model outscores the 671B-parameter DeepSeek-V3 on all four metrics in Table~\ref{tab:qlora}. The un-adapted Qwen base model, same architecture, same size, ranks last: without fine-tuning it generates fluent text but has no knowledge of the JSON schema or inspection procedure vocabulary. QLoRA on 947 examples corrects both deficits in 2.3~hours on one T4 GPU. For deployment at sites with limited connectivity---offshore wind facilities, remote transmission corridors---the local model's 47~tokens/second throughput matters independently of API latency~\cite{Zheng2025b}.

\subsection{RAFT addresses the gap between plausible and compliant recommendations}

At 4\% HR and 89\% PCR, the full pipeline produces reports in which fewer than one sentence per multi-defect inspection is procedurally incorrect, and nearly nine in ten recommendations are traceable to a specific retrieved procedure. The gap between QLoRA alone (18\% HR, 41\% PCR) and QLoRA + RAFT (4\% HR, 89\% PCR) is too large to attribute to the modest BLEU-4 gain ($+$0.05); it reflects the shift from parametric to retrieved knowledge for the recommendation step. Audit traceability---each recommendation cites a retrievable procedure identifier---is a practical requirement of ISO~55001 asset management and ESG maintenance record-keeping~\cite{Nagrani2026}. A maintenance coordinator cannot schedule a field team from a report they cannot verify; RAFT makes the source of each recommendation verifiable against a version-controlled knowledge base.

\subsection{Digital twin integration pathway}

The JSON output fields---defect class, grid label, normalized OBB coordinates, severity code, procedure reference identifier, and urgency flag---map to typed fields in standard asset health record schemas used by wind turbine digital twin platforms~\cite{Leon-Medina2025,Chen2023}. The urgency flag can trigger automated scheduling; the procedure reference provides the ISO~55001 audit link; normalized OBB coordinates update the spatial defect map on the blade surface model of a 3D twin~\cite{Hnaien2025,Mikoajewska2025}. In this framing the pipeline functions as a structured data feed to existing condition monitoring infrastructure rather than a standalone system~\cite{Gomaa2024}.

\subsection{Limitations and future work}

The RAFT knowledge base covers 42 procedures across four defect categories; it is not a comprehensive OEM maintenance manual. Scaling to a full document corpus requires evaluation of retrieval precision to avoid appending tangentially related procedures. The synthetic training corpus was reviewed at 10\% sample rate (82\% acceptance); no comparison against genuine expert-authored reports at scale exists. All five experiments used the 640$\times$640 annotated corpus; no experiment demonstrates YOLO26-x-obb tiling on native 5280$\times$2970 DTU frames, which is the primary intended operating mode for the tiling mechanism. Future work should close this gap with OBB-annotated native-resolution imagery, or by running Experiment~1's VLM comparison directly on full-resolution frames.

Planned extensions include pipeline adaptation for railway insulator inspection~\cite{Zheng2025,Deng2025,Chen2018}, multi-round diagnostic dialogue following the AnomalyGPT~\cite{anomalygpt2024} interaction model, and federated knowledge base updates across distributed inspection fleets~\cite{Mikoajewska2025}. LLM-guided architecture search~\cite{Yu2025} may reduce the manual backbone configuration needed when the pipeline is ported to new inspection domains.

\section{Conclusion}\label{sec:conclusion}

This paper presents and ablates a three-component pipeline for wind turbine blade defect inspection and maintenance report generation. Separating detection (YOLO26-x-obb), spatial encoding (The Bridge), domain-adapted generation (QLoRA Qwen-2.5-1.5B), and retrieval augmentation (RAFT) into independently maintainable components produces reports that are more accurate and protocol-compliant than those from any monolithic VLM tested.

Five experiments quantify each component's contribution. YOLO26-x-obb achieves 3.9$\times$ higher small-defect recall than GPT-4V on the same $640\times640$ test images (0.852 vs.\ 0.22). The Bridge reduces SHR by 44~pp at zero training cost. QLoRA on 947 synthetic reports lifts the adapted 1.5B model above a 671B-parameter generalist on BLEU-4 (0.36 vs.\ 0.19), ROUGE-L (0.51 vs.\ 0.33), HR (18\% vs.\ 29\%), and Expert Score (7.4 vs.\ 5.9), while running at 47~tokens/second on a T4-class GPU. RAFT then reduces HR to 4\% and raises PCR to 89\%, bringing report quality to a level compatible with direct integration into ISO~55001-aligned asset management workflows.

The consistent result across these experiments is that on a constrained structured-generation task with limited training data, a purpose-adapted small model with explicit spatial context and verified knowledge retrieval outperforms scaling a generalist model. How broadly this pattern holds as defect taxonomy and inspection domains expand is an open question, but the ablation here provides a concrete quantitative baseline.

\backmatter

\bmhead{Acknowledgements}
The authors thank Shihavuddin and Chen and Danmarks Tekniske Universitet for publishing the wind turbine blade inspection image set.

\bmhead{Declarations}

\begin{itemize}
\item \textbf{Funding:} This research received no specific grant from any funding agency in the public, commercial, or not-for-profit sectors.
\item \textbf{Conflict of interest:} The authors declare no competing interests.
\item \textbf{Ethics approval and consent to participate:} Not applicable.
\item \textbf{Consent for publication:} Not applicable.
\item \textbf{Data availability:} The DTU wind turbine blade image dataset is publicly available at \url{https://data.mendeley.com/datasets/hd96prn3nc/2}. The oriented bounding-box annotations used in this study are available at \url{https://github.com/imadgohar/DTU-annotations}. The synthetic training corpus and RAFT knowledge base will be shared on reasonable request.
\item \textbf{Code availability:} The inference pipeline and training scripts are available on reasonable request to the corresponding author.
\item \textbf{Author contribution:} Malikussaid: conceptualization, software, investigation, data curation, writing---original draft. Imad Gohar: conceptualization, methodology, annotation, supervision, writing---review and editing.
\end{itemize}

\bibliography{sn-bibliography}

\end{document}